\documentclass[10pt,twocolumn,letterpaper]{article}

\usepackage{cvpr}              

\usepackage[dvipsnames]{xcolor}

\usepackage{graphicx}
\usepackage{amsmath}
\usepackage{amssymb}
\usepackage{booktabs}

\usepackage{xcolor} 
\usepackage{times}
\usepackage{epsfig}
\usepackage{graphicx}
\usepackage{amsmath}
\usepackage{amssymb}
\usepackage[bb=dsserif]{mathalpha}
\usepackage{bm}

\usepackage{booktabs,colortbl,tabularx}
\usepackage{pifont}%

\usepackage{mathtools}
\usepackage{commath}
\usepackage{algorithm}
\usepackage[noend]{algpseudocode}

\usepackage{multirow}
\usepackage{comment} 
\usepackage{enumitem}
\usepackage{minted}

\definecolor{Gray}{gray}{0.9}

\newcommand{\cmark}{\ding{51}}%
\newcommand{\xmark}{\ding{55}}%

\definecolor{battleshipgrey}{rgb}{0.52, 0.52, 0.51}

\usepackage{xspace}
\newcommand{\method}{{\fontfamily{lmtt}\selectfont\emph{\textsc{Foley-Flow}}}\xspace}

\definecolor{cvprblue}{rgb}{0.21,0.49,0.74}
\usepackage[pagebackref,breaklinks,colorlinks,allcolors=cvprblue]{hyperref}

\def\paperID{10802} 
\def\confName{CVPR}
\def\confYear{2025}

\title{Foley-Flow: Coordinated Video-to-Audio Generation with\\ Masked Audio-Visual Alignment and Dynamic Conditional Flows}

\author{%
  Shentong Mo\\
  CMU / MBZUAI\\
  DAMO Academy, Alibaba Group\\
  {\tt\small shentongmo@gmail.com} \\
   \and
   Yibing Song\thanks{Y. Song is the corresponding author.} \\
   DAMO Academy, Alibaba Group \\
   Hupan Laboratory \\
   {\tt\small songyibing.syb@alibaba-inc.com} \\
}

\begin{document}
\maketitle

\begin{abstract}

    Coordinated audio generation based on video inputs typically requires a strict audio-visual (AV) alignment, where both semantics and rhythmics of the generated audio segments shall correspond to those in the video frames.
    Previous studies leverage a two-stage design where the AV encoders are firstly aligned via contrastive learning, then the encoded video representations guide the audio generation process. We observe that both contrastive learning and global video guidance are effective in aligning overall AV semantics while limiting temporally rhythmic synchronization. 
    In this work, we propose \method to first align unimodal AV encoders via masked modeling training, where the masked audio segments are recovered under the guidance of the corresponding video segments. After training, the AV encoders which are separately pretrained using only unimodal data are aligned with semantic and rhythmic consistency. Then, we develop a dynamic conditional flow for the final audio generation. Built upon the efficient velocity flow generation framework, our dynamic conditional flow utilizes temporally varying video features as the dynamic condition to guide corresponding audio segment generations. To this end, we extract coherent semantic and rhythmic representations during masked AV alignment, and use this representation of video segments to guide audio generation temporally. Our audio results are evaluated on the standard benchmarks and largely surpass existing results under several metrics. The superior performance indicates that \method is effective in generating coordinated audios that are both semantically and rhythmically coherent to various video sequences.

\end{abstract}

\section{Introduction}

Video sequences are naturally captured with audio signal accompaniment. The audio reflects the sound of the object types and motions in the video~\cite{mo2022EZVSL,mo2022SLAVC,mo2022benchmarking,mo2022semantic,mo2022multimodal,mo2023avsam,mo2023weaklysupervised,mo2023oneavm}. For one video where the accompanied audio is missing, coordinated audio generation supplements the video stream to formulate complete human perception. Generating coordinated audio from video inputs is typically from two perspectives. First, the audio should reflect the unique semantics of the video content (\textit{e.g.}, the sound of a bird chirping is different from the sound of a dog barking). Second, the sound in one audio should be coherent with the object motions in the video sequence (\textit{e.g.}, the clip-clop of the horse's hooves corresponds to the horse walking on the road. The clip-clop happens only when the hooves hit the ground). As such, the semantics and rhythmics of the generated audio segments should correspond to the video contents. 

\begin{figure}[t]
\centering
\includegraphics[width=\linewidth]{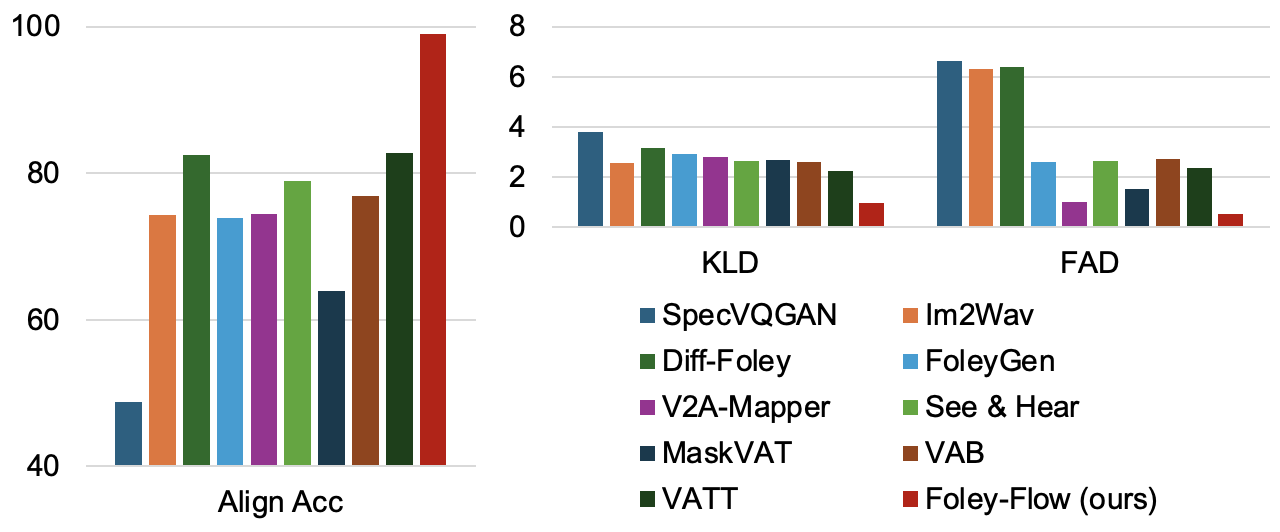}
\vspace{-0.5em}
\caption{Comparison with state-of-the-art approaches. 
We achieve the best results regarding semantical (Align Acc) and rhythmical (KLD, FAD) metrics.}
\label{fig: title_img}
\vspace{-0.5em}
\end{figure}

Existing methods~\cite{Iashin2021SpecVQGAN, sheffer2023im2wav, kreuk2023audiogen, luo2023difffoley} on video-to-audio generation can be summarized as a two-step design. In the first step, a visual encoder (\textit{e.g.}, a ViT) is aligned to an audio encoder (\textit{i.e.}, a ViT takes audio spectrograms). The visual/audio encoder is pretrained using unimodal data only, with a self-supervised or supervised pretraining scheme. For AV encoder alignment, contrastive learning across AV data pairs is typically introduced to formulate unified AV representations. In the second step, an audio generation framework is utilized where the initial audio noise is gradually refined to the final audio signal. During this progress, video representations are involved as the condition for audio generation. As the video representations are from the visual encoder that is aligned to the audio in the first step, the condition of video content is effective in guiding audio generation. 

The contrastive learning for AV alignment~\cite{mo2023audiovisual,mo2023classincremental,mo2023deepavfusion,mo2024audiovisual,mo2024semantic,pian2023audiovisual,mahmud2024maavt,mo2024aligning} and video-guided audio generation~\cite{mo2023diffava,mo2024texttoaudio,zhang2024audiosynchronized,zhang2025scaling}, although benefiting global AV semantic alignment, limits their performance upon local rhythmic correspondence. This is because contrastive learning regards the video and audio pairs as a whole, while the temporal segments within each AV data pair are not differentiated. Also, the video data is typically utilized as a whole sample to guide the audio generation, without specifying the sound appearance (\textit{i.e.}, rhythm and amplitude) in each audio segment. Indeed, coordinated video-to-audio generation should capture both high-level semantic cues (\textit{e.g.}, the type of sound associated with visual contents) and low-level rhythmic patterns that correspond with the timing of events within the video. These challenges limit existing methods to produce natural and synchronized audio, where the segments within AV data pairs shall be processed individually rather than in a general form.

In this work, we address these challenges by proposing \method, to advance the AV correspondence from the temporal segment level during both alignment and generation steps. Given one visual encoder and one audio encoder, we propose a masked AV alignment scheme where the audio encoder is learned to recover the masked audio segments based on the temporally sequential video segments. Our masked AV alignment enables representations to contain both semantics and synchronization, which is the foundation of the final audio generation. In the next step, we propose a dynamic conditional flow that is built upon the velocity flow based generation framework. Our conditions can be updated dynamically in accordance with the temporal varying video segments. To this end, the semantics and rhythmics of the video representations, aligned with audio representations through the first step, are fully explored for the flow-based audio generations. The temporally varying video features sequentially guide corresponding audio segment generation step-by-step via the efficient inference of our recurrent flow-based model. We validate our \method on the VGGSound dataset, a comprehensive collection of videos with diverse audio-visual content. Experimental results demonstrate that \method achieves state-of-the-art performance across key metrics, including Frechet Audio Distance (FAD), Kullback-Leibler (KL) divergence, and alignment accuracy. These metrics underscore our model's ability to produce synchronized and contextually accurate audio outputs, setting a new benchmark for video-to-audio generation. 

We summarize our contributions below:
\begin{itemize}
    \item We introduce \method\ with masked audio-video alignment to enhance semantic and rhythmic correspondence in audio-visual data pairs.
    \item We propose a dynamic conditional flow to generate final audio results based on the temporally varying video representations step-by-step.
    \item We demonstrate the effectiveness of our approach through extensive evaluations in achieving state-of-the-art performance across multiple audio quality metrics.
\end{itemize}

\section{Related Works}

\subsection{Video-to-Audio Generation}
Video-to-audio generation, the task of translating visual information into corresponding audio outputs, has seen rapid advancements driven by innovative model architectures.  
SpecVQGAN~\cite{Iashin2021SpecVQGAN} employs vector quantized generative adversarial networks to transform visual features into audio spectrograms. Im2Wav~\cite{sheffer2023im2wav} leverages CLIP embeddings to generate audio waveforms directly from images. Diff-Foley~\cite{luo2023difffoley} aligns video and audio embeddings using Contrastive Audio-Visual Pre-training (CAVP) and refines the results using diffusion models. FoleyGen~\cite{mei2023foleygen} adopts a language modeling approach combined with a neural audio codec for waveform-to-token conversion.  

Recent works have explored multimodal pre-training and fine-tuning strategies. Seeing \& Hearing~\cite{xing2024seeing} utilizes the pre-trained ImageBind model~\cite{girdhar2023imagebind} as a latent aligner for cross-modal diffusion-based generation. VAB~\cite{su2024vision} introduces a visual-conditioned masked audio token prediction task for pre-training without diffusion models. MaskVAT~\cite{pascual2024masked} combines a sequence-to-sequence masked generative model with a high-quality audio codec to achieve temporal synchronicity. VATT~\cite{liu2024tell} integrates large language models like Gemma-2B~\cite{gemmateam2024gemma} and LLama-2-7B~\cite{touvron2023llama} with projection layers to map video features into audio tokens. 
While these methods address semantic alignment, they often neglect rhythmic synchronization, leading to asynchronous audio generation. 
Our approach bridges this gap by explicitly modeling both semantic and rhythmic coherence through video-audio masking alignment and generalized video-audio flow.

\subsection{Audio-Visual Masked Autoencoders}
Masked modeling has proven effective for learning audio-visual representations, with diverse pipelines proposed in prior work~\cite{huang2022mavil,gong2023contrastive}. For instance, CAV-MAE~\cite{gong2023contrastive} combines contrastive learning with masked modeling to fuse cross-modal information through a joint encoder. MAViL~\cite{huang2022mavil} extends this approach by introducing intra- and inter-modal contrastive learning along with masked reconstruction to capture contextualized representations. MBT~\cite{nagrani2022attention} enforces cross-modal information sharing through bottleneck latents to improve representation learning.

In contrast to these methods, our framework directly leverages masked modeling for video-to-audio generation. By masking audio segments and tasking the model with reconstruction based on video features, our video-audio masking alignment module enables the learning of synchronization patterns and precise temporal alignment. This design goes beyond representation learning to address the challenges of generation quality and rhythmic coherence in video-to-audio tasks.
Our \method\ advances the state of the art by integrating innovations from diffusion models, video-to-audio generation, and masked autoencoders into a unified framework. 
Unlike existing methods, the proposed \method\ achieves both semantic and rhythmic alignment through the video-audio masking alignment module and enhances generation efficiency using the generalized video-audio flow module, setting new benchmarks for video-to-audio generation tasks.

\subsection{Diffusion Models}
Diffusion models have emerged as a powerful paradigm for generative tasks across multiple domains. The foundational works on denoising diffusion probabilistic models (DDPMs)~\cite{ho2020denoising,song2021scorebased} introduced a framework that iteratively corrupts data with Gaussian noise in a forward process and trains a model to reverse this process, recovering the original data. These models have demonstrated remarkable success in applications such as image generation~\cite{saharia2022photorealistic,mo2024scaling}, image restoration~\cite{saharia2021image}, speech synthesis~\cite{kong2021diffwave}, point clouds~\cite{mo2023dit3d,mo2023fastdit3d,mo2024efficient}, and video generation~\cite{ho2022imagen}.  
Building upon this foundation, our work integrates a flow-based objective into the video-to-audio generation pipeline to accelerate inference and enhance output quality, addressing the limitations of traditional diffusion-based methods in efficiency and scalability.

\section{Proposed Method}

In this section, we present the proposed coordinated video-to-audio generation framework, namely \method, which enables semantic and rhythmic coherence by leveraging masked audio-visual modeling and generalized flows. 
We first provide preliminaries in Section~\ref{sec: pre}, then present the Video-Audio Masking Alignment in Section~\ref{sec: vama} to learn semantic and rhythmic coherence, and finally introduce Generalized Video-Audio Flow in Section~\ref{sec: gvaf} to accelerate the video-audio denoising process for fast video-to-audio generation.

\subsection{Preliminaries}\label{sec: pre}

In this section, we first describe the problem setup and notations, and then revisit the contrastive video-audio pre-training and the denoising diffusion probabilistic models.

{\flushleft \bf Problem Setup and Notations.}
The task of video-to-audio generation involves synthesizing temporally aligned audio $\mathbf{A}$ from sequential video frames $\mathbf{V}$. Let $\mathbf{F}^v \in \mathbb{R}^{T_v \times D_v}$ and $\mathbf{F}^a \in \mathbb{R}^{T_a \times D_a}$ denote the features extracted from the video and audio modalities, respectively, where $T_v$ and $T_a$ are the temporal lengths and $D_v$, $D_a$ are the feature dimensions. The goal is to ensure that the generated audio $\mathbf{A}$ is both semantically coherent and rhythmically aligned with $\mathbf{V}$.

{\flushleft \bf Masked Autoencoders.}  
Generalized Masked Autoencoders~\cite{he2021masked} have demonstrated their effectiveness in unimodal representation learning. In the context of video and audio, unimodal masking involves randomly masking a subset of input tokens $\mathbf{X}_{\text{mask}}$ and reconstructing them using the remaining unmasked tokens $\mathbf{X}_{\text{unmask}}$. For video frames $\mathbf{V}$ and audio spectrograms $\mathbf{A}$, the reconstruction objective is:
\begin{equation}
    \mathcal{L}_{\text{recon}} = \|\mathbf{X}_{\text{mask}} - \hat{\mathbf{X}}_{\text{mask}}(\mathbf{X}_{\text{unmask}})\|^2.
\end{equation}
This self-supervised approach enables the model to learn rich unimodal representations.

{\flushleft \bf Flow-Based Generations.} 
Flow-based generative models learn invertible mappings from simple noise distributions to complex target distributions. Given a latent variable $\mathbf{z}$, the flow transformation $f_{\boldsymbol{\phi}}$ maps $\mathbf{z}$ to the target distribution:
\begin{equation}
    \mathbf{x} = f_{\boldsymbol{\phi}}(\mathbf{z}),
\end{equation}
where $\mathbf{z} \sim \mathcal{N}(\mathbf{0}, \mathbf{I})$. The objective minimizes the Kullback-Leibler divergence between the generated and target distributions:
\begin{equation}
    \mathcal{L}_{\text{flow}} = D_{\text{KL}}(p(\mathbf{x}) \| p_{\text{target}}(\mathbf{x})).
\end{equation}
In \method, we develop a dynamic conditional flow where the conditions are temporally varying video features.

\begin{figure}[t]
\centering
\includegraphics[width=0.95\linewidth]{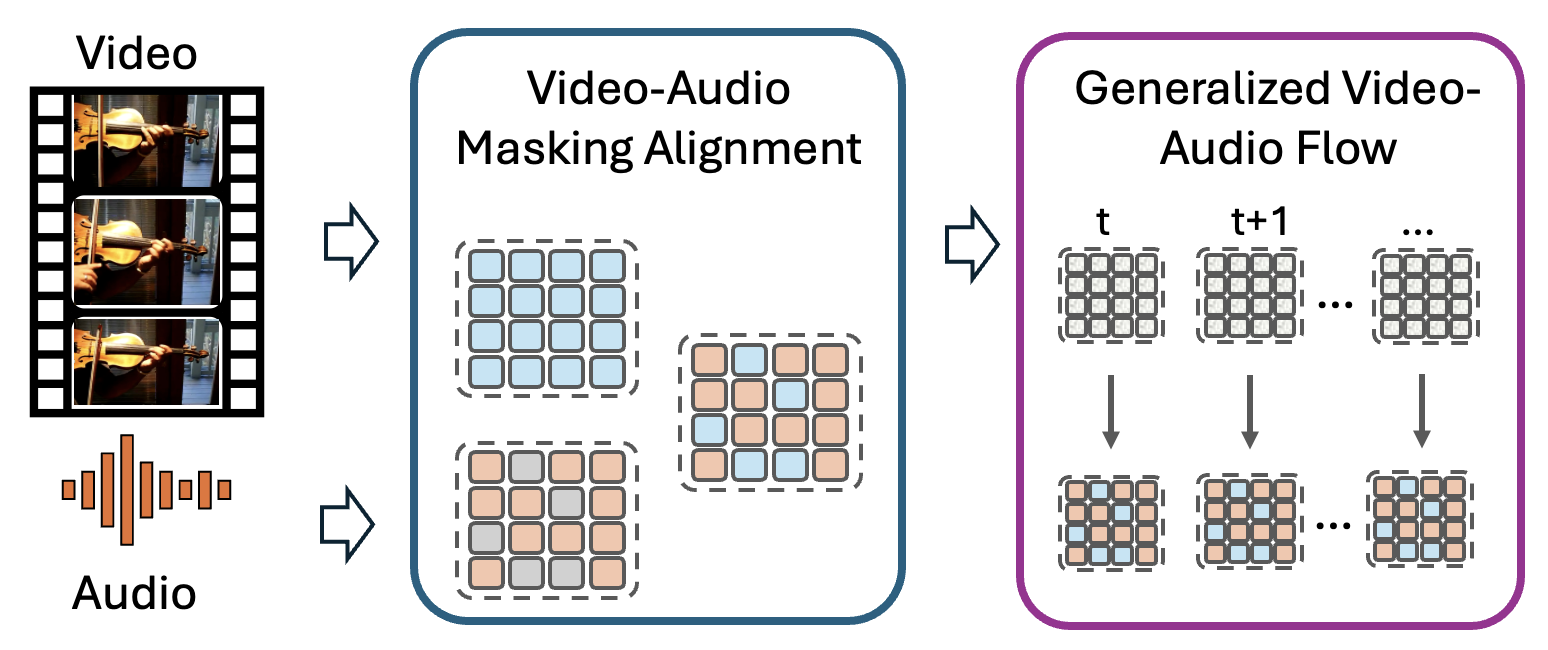}
\vspace{-0.5em}
\caption{Illustration of the proposed \method\ framework.
}
\label{fig: title_img}
\vspace{-0.5em}
\end{figure}

\subsection{Video-Audio Masking Alignment}\label{sec: vama}

Traditional approaches to video-to-audio generation often rely on contrastive learning to align audio-visual representations. While effective for global semantic alignment, these methods struggle to capture rhythmic synchronization, as they do not explicitly model the temporal dependencies between audio and video segments. To address this limitation, we propose a Video-Audio Masking Alignment (VAMA) strategy that forces the model to reconstruct masked audio segments using temporally corresponding video features. This not only aligns audio and video representations semantically but also ensures that their temporal rhythms are synchronized.

To ensure both semantic and rhythmic coherence, we introduce a novel cross-modal masking strategy that leverages the complementary nature of video and audio information. This strategy involves masking a subset of audio segments $\mathbf{F}^a_{\text{mask}}$ and tasking the model with reconstructing these segments using frame-wise video features $\mathbf{F}^v$ along with the unmasked audio inputs $\mathbf{F}^a_{\text{unmask}}$. 
This setup mimics real-world scenarios where incomplete or noisy audio must be inferred from video cues and the remaining audio context.
The self-supervised reconstruction objective is defined as:
\begin{equation}
    \mathcal{L}_{\text{VAMA}} = \|\mathbf{F}^a_{\text{mask}} - \hat{\mathbf{F}}^a_{\text{mask}}(\mathbf{F}^v, \mathbf{F}^a_{\text{unmask}})\|^2,
\end{equation}
where $\hat{\mathbf{F}}^a_{\text{mask}}$ denotes the model’s reconstructed audio segments. The masking is performed temporally, ensuring that the model learns to capture the synchronization patterns between video frames and their corresponding audio segments.

To achieve this, the video encoder extracts temporal features $\mathbf{F}^v$ that encapsulate spatial and contextual information from the video frames. These features are aligned with audio features $\mathbf{F}^a$ in the latent space through a cross-attention mechanism, which ensures that the temporal dynamics of audio are synchronized with video features. The masking strategy encourages the model to focus on rhythmic patterns, as the temporal continuity of masked audio segments must be reconstructed using visual and audio cues.
By combining cross-modal masking with temporal consistency, the Video-Audio Masking Alignment module enables the model to learn robust representations for both semantic and rhythmic coherence.

Unlike traditional contrastive methods that only align global representations, our video-audio masking alignment explicitly aligns local and temporal features. This approach enforces rhythmic consistency between modalities, which is critical for synchronized video-to-audio generation. The novel use of segment-level masking in a cross-modal context also introduces a new way to train audio-visual models for temporal alignment.

\subsection{Generalized Video-Audio Flow}\label{sec: gvaf}

Flow-based models are known for their ability to generate high-quality outputs with efficient sampling. However, existing flow-based methods in generative tasks often use static conditions, which are insufficient for capturing the dynamic nature of video-to-audio generation. Video features evolve over time, reflecting changes in scene content, motion, and rhythm, which should be dynamically integrated into the audio generation process. 
To address this, we propose a dynamic conditional flow module that conditions the audio generation on temporally varying video segments, ensuring that the generated audio maintains both semantic and rhythmic alignment.

{\flushleft \bf Dynamic Conditional Flow.}
Our flow-based model builds upon the velocity flow framework, incorporating temporally varying video features $\mathbf{F}^v_t$ as dynamic conditions. The flow transformation is defined as:
\begin{equation}
    \mathbf{F}^a_t = f_{\boldsymbol{\phi}}(\mathbf{z}_t, \mathbf{F}^v_t),
\end{equation}
where $\mathbf{z}_t$ is a noise vector sampled from a Gaussian distribution. Unlike static flows, the inclusion of temporal conditions allows the model to adapt the generated audio to reflect changes in video dynamics over time.

To effectively model temporal synchronization, we use video segment features $\mathbf{F}^v_t$ as conditions for the flow-based generator. These features capture both high-level semantics (e.g., object actions) and low-level rhythms (e.g., motion patterns) that vary across video frames. By conditioning the flow model on these dynamic features, the generated audio $\mathbf{F}^a_t$ evolves in a temporally coherent manner, ensuring alignment with the video’s temporal structure.

The dynamic conditional flow module introduces a novel way of incorporating temporally varying conditions into flow-based generative models. Unlike static conditioning approaches, our method dynamically adapts the audio generation process to reflect video segment changes, achieving fine-grained synchronization. This temporal adaptability is particularly suited for video-to-audio tasks, where the rhythm and semantics of the audio must closely follow the video’s progression.

{\flushleft \bf Generalized Dynamic Conditional Flow.}
To overcome the inefficiencies of traditional denoising processes and enable fast and high-quality video-to-audio generation, we introduce a generalized dynamic conditional flow module. This module leverages normalizing flows to provide a continuous, invertible mapping from simple noise distributions to complex target audio distributions. The flow-based approach not only improves inference speed but also enhances the model's ability to generate high-fidelity audio outputs.

{\flushleft \bf Flow-based Mapping.}  
Given a latent representation of the video $\mathbf{z}_v$, the flow-based generator learns a transformation $f_{\boldsymbol{\phi}}$ that maps $\mathbf{z}_v$ and a noise vector $\mathbf{z}_{\text{noise}} \sim \mathcal{N}(\mathbf{0}, \mathbf{I})$ to the corresponding audio latent $\mathbf{z}_a$:
\begin{equation}
    \mathbf{z}_a = f_{\boldsymbol{\phi}}(\mathbf{z}_v, \mathbf{z}_{\text{noise}}).
\end{equation}
The invertibility of $f_{\boldsymbol{\phi}}$ ensures that the generated audio can be effectively traced back to its original latent distribution, enabling precise alignment with the video features.

{\flushleft \bf Training Objective.}  
The Generalized Video-Audio Flow (GVAF) module is trained to minimize the Kullback-Leibler divergence between the distribution of generated audio features $p(\mathbf{z}_a)$ and the target distribution $p_{\text{target}}(\mathbf{z}_a)$:
\begin{equation}
    \mathcal{L}_{\text{GVAF}} = D_{\text{KL}}(p(\mathbf{z}_a) \| p_{\text{target}}(\mathbf{z}_a)).
\end{equation}
This objective ensures that the generated audio closely matches the target audio in both distributional and temporal characteristics.

{\flushleft \bf Efficient Inference with Conditional Flows.}  
During inference, the flow-based module accelerates the video-to-audio generation process by reducing the number of iterative steps required for synthesis. Instead of relying on a time-consuming denoising process, the conditional flow directly generates high-quality audio in a single step:
\begin{equation}
    \mathbf{z}_a = f_{\boldsymbol{\phi}}^{-1}(\mathbf{z}_v, \mathbf{z}_{\text{noise}}).
\end{equation}
This approach is computationally efficient and well-suited for real-time applications, as it bypasses the iterative nature of traditional diffusion models.

{\flushleft Integration with Masking Alignment.}  
To further enhance the generation quality, the flow-based module is conditioned on the features learned from the Video-Audio Masking Alignment module. This conditioning provides the flow with semantically rich and rhythmically coherent video representations, ensuring the generated audio is contextually aligned with the visual input.

By integrating the flow-based module with the masking alignment strategy, our \method achieves a balance between efficiency and quality, producing synchronized, high-fidelity audio outputs in real-time settings.

\section{Experiments}

\subsection{Experimental Setup}

\noindent \textbf{Datasets.}
We conduct our experiments on two widely used datasets:  
VGGSound~\cite{chen2020vggsound} consists of 200k YouTube video clips, each 10 seconds long, spanning 309 diverse sound categories such as animals, vehicles, human speech, dancing, and musical instruments.  
AudioSet~\cite{Gemmeke2017audioset} is a larger audio-visual database containing approximately 2M YouTube videos with a wide variety of audio-visual content.

\noindent \textbf{Evaluation Metrics.}
To evaluate video-to-audio generation quality, we adopt the following metrics: 
Kullback-Leibler Divergence (KLD) measures how closely the generated audio matches the ground truth (GT) by comparing the distributions of PaSST features~\cite{koutini2022efficient}, capturing how well the audio represents the concepts in the video.  
Fréchet Audio Distance (FAD)~\cite{kilgour2018frechet} assesses the distributional quality of the generated audio compared to real audio, providing a measure of overall quality.  
Alignment Accuracy (Align Acc)~\cite{luo2023difffoley} evaluates the temporal alignment and relevance between generated audio and corresponding video frames.  
Generation Speed measures efficiency by computing the time taken to generate a waveform sample.

\noindent \textbf{Implementation.}
The input video frames are resized to $224 \times 224$ resolution. Audio inputs are represented as log spectrograms extracted from 10-second clips sampled at 8000Hz. Following prior work~\cite{mo2022EZVSL}, the audio spectrograms are generated using a Short-Time Fourier Transform (STFT) with 50ms windows and a 25ms hop size, resulting in input tensors of size $128 \times 128$ (128 frequency bands over 128 timesteps).  
For the audio and visual encoder, we use the single-modality encoders~\cite{sun2023evaclip,huang2022amae}  to initialize the visual encoder using weights pre-trained on LAION-400M~\cite{schuhmann2021laion400m} and audio encoder pre-trained on AudioSet~\cite{Gemmeke2017audioset}.
The models were trained for 100 epochs using the Adam optimizer~\cite{kingma2014adam} with a learning rate of $1e-4$ and a batch size of $128$.

\begin{table}[t]
\centering
\caption{Comparison results on VGGSound test set for video-to-audio generation. The KLD, FAD, and Align Acc values are reported.}
\label{tab: exp_sota}
\scalebox{0.86}{
\begin{tabular}{lccc}
\toprule
Method            & KLD $\downarrow$ & FAD $\downarrow$ & Align Acc $\uparrow$ \\ \midrule
SpecVQGAN~\cite{Iashin2021SpecVQGAN}         & 3.78 & 6.63 & 48.79 \\
Im2Wav~\cite{sheffer2023im2wav}            & 2.54 & 6.32 & 74.31 \\
Diff-Foley~\cite{luo2023difffoley}        & 3.15 & 6.40 & 82.47 \\
FoleyGen~\cite{mei2023foleygen}          & 2.89 & 2.59 & 73.83 \\
V2A-Mapper~\cite{wang2024v2amapper}        & 2.78 & 0.99 & 74.37 \\
Seeing \& Hearing~\cite{xing2024seeing} & 2.62 & 2.63 & 78.95 \\
MaskVAT~\cite{pascual2024masked}           & 2.65 & 1.51 & 63.87 \\
VAB~\cite{su2024vision}               & 2.58 & 2.69 & 76.83 \\
VATT~\cite{liu2024tell}              & 2.25 & 2.35 & 82.81 \\
\method\ (ours) & \bf 0.97 & \bf 0.52 & \bf 98.97 \\
\bottomrule
\end{tabular}}
\end{table}

\subsection{Comparison to Prior Work}\label{sec:exp}

In this work, we propose a novel and effective framework called \method, for video-to-audio generation. 
In order to demonstrate the effectiveness of the proposed \method, we conduct an extensive evaluation of its performance against a wide range of established video-to-audio generation models.
SpecVQGAN (BMVC 2021)~\cite{Iashin2021SpecVQGAN} is a vector quantized GAN for audio spectrogram generation;
Im2Wav (ICASSP 2023)~\cite{sheffer2023im2wav} uses CLIP embeddings to directly generate waveforms from images;  
Diff-Foley (NeurIPS 2023)~\cite{luo2023difffoley} combines contrastive pre-training with diffusion models for audio generation;
FoleyGen (arXiv 2023)~\cite{mei2023foleygen} utilizes a neural audio codec and a language modeling paradigm for bidirectional waveform-to-token conversion;
Seeing \& Hearing (CVPR 2024)~\cite{xing2024seeing} employs a multimodal latent aligner guided by ImageBind~\cite{girdhar2023imagebind} for diffusion-based cross-modal generation; 
VAB (ICML 2024)~\cite{su2024vision} pre-trains visual-conditioned masked audio token prediction for representation learning without diffusion; 
MaskVAT (ECCV 2024)~\cite{pascual2024masked} combines a high-quality general audio codec with a masked generative model for semantic and temporal alignment;  
VATT (NeurIPS 2024)~\cite{liu2024tell} leverages LLMs fine-tuned for instruction to map video features into the audio token space using parallel decoding.

Table~\ref{tab: exp_sota} reports a comprehensive comparison of \method\ against state-of-the-art video-to-audio generation models on the VGGSound test set. 
Lower KLD and FAD values indicate better semantic and overall audio quality, while higher Align Acc reflects superior temporal synchronization between the video and the generated audio.
KLD measures the semantic similarity between the generated and ground-truth audio distributions. \method\ achieves the lowest KLD score of 0.97, significantly outperforming the next best model, VATT, with a KLD of 2.25. This result demonstrates that \method\ excels in capturing semantic consistency in the generated audio, aligning it closely with the content of the video frames. The improvement over models like Diff-Foley (3.15) and MaskVAT (2.65) highlights the effectiveness of our Video-Audio Masking Alignment module in aligning semantic representations during training.
FAD evaluates the overall quality and distributional similarity of the generated audio to real audio. \method\ achieves an FAD of 0.52, the best among all models. This is a substantial improvement over the previous best-performing model, FoleyGen, which has an FAD of 2.59. The results indicate that the dynamic conditional flow used in \method\ not only generates high-quality audio but also maintains coherence across temporal segments, ensuring smooth transitions and realistic audio outputs.
Align Acc measures the temporal synchronization between the video and generated audio. \method\ achieves an Align Acc of 98.97, setting a new benchmark and significantly surpassing the closest competitor, Diff-Foley, which achieves an Align Acc of 82.47. This improvement underscores the ability of \method\ to model rhythmic dependencies effectively, leveraging dynamic video conditions to guide audio generation. The incorporation of temporally evolving video features in the flow-based generation process ensures that the generated audio matches the timing of events in the video.

SpecVQGAN exhibits poor performance across all metrics, with high KLD (3.78) and FAD (6.63) and low Align Acc (48.79). This highlights the limitations of generative adversarial networks in capturing cross-modal dependencies.
Im2Wav and V2A-Mapper improve upon SpecVQGAN but fall short in rhythmic synchronization, with Align Acc scores of 74.31 and 74.37, respectively.
While Diff-Foley shows strong performance in Align Acc (82.47), its high KLD (3.15) and FAD (6.40) indicate suboptimal audio quality and semantic alignment.
FoleyGen, Seeing \& Hearing, and VAB demonstrate balanced performance but fail to match the semantic and rhythmic alignment achieved by \method, as indicated by their lower Align Acc and higher KLD scores.
Although MaskVAT and VATT utilize advanced audio-visual encoding strategies, their static conditions and lack of temporal adaptability lead to inferior results compared to \method.
The results demonstrate that \method\ significantly outperforms state-of-the-art methods in all three metrics. The integration of the Video-Audio Masking Alignment module and the Dynamic Conditional Flow module allows \method\ to generate audio that is semantically accurate, temporally synchronized, and of high quality. These advancements make \method\ a robust framework for video-to-audio generation, setting new standards for performance in this domain.

We provide qualitative results in the supplementary material comparing \method\ against baseline methods. The generated audio outputs of \method\ are semantically richer, rhythmically synchronized, and more temporally aligned with the video content. Notably, \method\ handles complex scenarios, such as delayed auditory responses to visual actions, with higher fidelity and naturalness than competing approaches.

\subsection{Experimental Analysis}

In this section, we performed detailed ablation studies to demonstrate the benefit of introducing the Video-Audio Masking Alignment (VAMA) and  Generalized Video-Audio Flow (GVAF) on our proposed framework. 
Furthermore, we conducted extensive experiments to explore the impact of the choice of video-audio encoders, video-audio flow objectives, and the effect of the masking ratio.

\begin{table}[t]
\centering
\caption{Ablation studies on Video-Audio Masking Alignment (VAMA) and Generalized Video-Audio Flow (GVAF) for video-to-audio generation. The KLD, FAD, and Align Acc values are reported.}
\label{tab: ab_component}
\scalebox{0.76}{
\begin{tabular}{ccccc}
\toprule
VAMA & GVAF  & KLD $\downarrow$ & FAD $\downarrow$ & Align Acc $\uparrow$ \\ \midrule
\xmark & \xmark & 3.15 & 6.40 & 82.47 \\
\cmark & \xmark & 1.68 & 1.57 & 96.29 \\
\xmark & \cmark & 1.92 & 1.89 & 93.86 \\
\cmark & \cmark & \bf 0.97 & \bf 0.52 & \bf 98.97\\
\bottomrule
\end{tabular}}
\end{table}

\noindent\textbf{Video-Audio Masking Alignment \& Generalized Video-Audio Flow.}
To evaluate the contributions of the Video-Audio Masking Alignment (VAMA) and Generalized Video-Audio Flow (GVAF) modules to \method, we conduct an ablation study by removing these components individually and in combination. The quantitative results, as reported in Table~\ref{tab: ab_component}, demonstrate the critical role of both modules in achieving state-of-the-art performance in video-to-audio generation.
When VAMA is removed, the Align Acc metric drops from \textit{98.97\%} to \textit{93.86\%}, highlighting a reduction in temporal synchronization between the generated audio and video frames. Additionally, the KLD increases from \textit{0.97} to \textit{1.92}, indicating that the semantic alignment between video and audio is less robust. These results underscore that VAMA's cross-modal masking strategy is essential for enforcing rhythmic synchronization and ensuring that the semantic representations of audio and video align effectively.
When GVAF is removed, the FAD score increases significantly from \textit{0.52} to \textit{1.57}, indicating a substantial decline in the quality of the generated audio. This is because the absence of GVAF prevents the model from leveraging temporally dynamic video conditions during the audio generation process. The Align Acc also drops slightly to \textit{96.29\%}, reflecting a weaker temporal alignment. These results highlight the role of GVAF in ensuring efficient and high-fidelity audio generation, particularly through its dynamic conditional flow mechanism.
When both modules are removed, the performance deteriorates significantly across all metrics. The KLD increases to \textit{3.15}, FAD rises to \textit{6.40}, and Align Acc plummets to \textit{82.47\%}. This dramatic decline indicates that both VAMA and GVAF are indispensable for achieving semantic coherence, rhythmic synchronization, and high-quality audio outputs. Without these modules, the model struggles to effectively align and generate audio that corresponds to the video inputs.
With both VAMA and GVAF included, \method\ achieves the best performance across all metrics: a KLD of \textit{0.97}, FAD of \textit{0.52}, and Align Acc of \textit{98.97\%}. These results validate the synergy between the two modules in addressing both semantic and rhythmic challenges in video-to-audio generation. VAMA ensures robust cross-modal alignment during training, while GVAF enables efficient and high-quality generation through its temporally adaptive flow-based approach.
The ablation study confirms the necessity of both VAMA and GVAF in \method. VAMA ensures semantic and rhythmic coherence through its cross-modal masking strategy, while GVAF provides dynamic temporal conditioning to enhance the quality and efficiency of audio generation. Together, these modules enable \method\ to outperform existing methods in video-to-audio generation.

\begin{table}[t]
\centering
\caption{Ablation studies on video-audio encoders for video-to-audio generation. The KLD, FAD, and Align Acc are reported.}
\label{tab: ab_encoder}
\scalebox{0.76}{
\begin{tabular}{ccccc}
\toprule
Video & Audio  & KLD $\downarrow$ & FAD $\downarrow$ & Align Acc $\uparrow$ \\ \midrule
ImageBind & ImageBind & 1.38 & 1.27 & 97.29 \\
MAE       & ImageBind & 1.17 & 1.06 & 98.32 \\
EVA-CLIP  & ImageBind & 1.01 & 0.73 & 98.52 \\
ImageBind & AudioMAE  & 1.36 & 1.13 & 97.74 \\
MAE       & AudioMAE  & 1.06 & 0.65 & 98.62 \\
EVA-CLIP  & AudioMAE  & \bf 0.97 & \bf 0.52 & \bf 98.97 \\
\bottomrule
\end{tabular}}
\end{table}

\noindent\textit{Type of Video-Audio Encoders.}
To investigate the impact of different encoder combinations for video and audio inputs, we evaluate various configurations using ImageBind~\cite{girdhar2023imagebind}, MAE~\cite{he2021masked}, and EVA-CLIP~\cite{sun2023evaclip} for video encoding, and ImageBind and AudioMAE~\cite{huang2022amae} for audio encoding. The quantitative results are reported in Table~\ref{tab: ab_encoder}.
Among the three video encoders, EVA-CLIP consistently achieves the best results across all metrics when paired with either audio encoder. For instance, when combined with AudioMAE, EVA-CLIP achieves the lowest KLD (\textit{0.97}), lowest FAD (\textit{0.52}), and highest Align Acc (\textit{98.97\%}). This demonstrates EVA-CLIP's ability to capture rich semantic representations and robust temporal features, which are essential for aligning video frames with corresponding audio segments.  
In contrast, ImageBind and MAE exhibit slightly lower performance. For example, ImageBind paired with AudioMAE results in a higher KLD of \textit{1.36} and lower Align Acc of \textit{97.74\%}, indicating weaker semantic and rhythmic alignment.
AudioMAE outperforms ImageBind in all cases, showcasing its strength in learning detailed audio representations. For instance, when paired with EVA-CLIP, AudioMAE achieves an FAD of \textit{0.52}, compared to \textit{0.73} when using ImageBind as the audio encoder. This suggests that AudioMAE is better suited for generating high-fidelity audio that aligns temporally with the video.
The combination of EVA-CLIP for video encoding and AudioMAE for audio encoding achieves the best results across all metrics, with a KLD of \textit{0.97}, FAD of \textit{0.52}, and Align Acc of \textit{98.97\%}. This indicates that EVA-CLIP’s pretraining on large-scale image-text datasets and AudioMAE’s focus on detailed spectrogram reconstruction are complementary, enabling superior semantic and rhythmic alignment in the generated audio.
These results highlight the importance of choosing robust encoder combinations for video-to-audio generation. The EVA-CLIP and AudioMAE pair provides the best balance between semantic richness and temporal alignment, leading to state-of-the-art performance.

\begin{table}[t]
\centering
\caption{Ablation studies on Dynamic Conditional Flow (DCF) and Generalized Dynamic Conditional Flow (GDCF) for video-to-audio generation. The KLD, FAD, and Align Acc are reported.}
\label{tab: ab_flow}
\scalebox{0.76}{
\begin{tabular}{cccc}
\toprule
Flow Type & KLD $\downarrow$ & FAD $\downarrow$ & Align Acc $\uparrow$ \\ \midrule
DCF  & 1.28 & 0.85 & 97.73 \\
GDCF & \bf 0.97 & \bf 0.52 & \bf 98.97 \\
\bottomrule
\end{tabular}}
\end{table}

\noindent\textbf{Type of Video-Audio Flow.}
To evaluate the impact of different flow objectives, we compare the performance of Dynamic Conditional Flow (DCF) and Generalized Dynamic Conditional Flow (GDCF) in the video-audio generation task. The quantitative results are summarized in Table~\ref{tab: ab_flow}.
The DCF approach achieves strong performance with a KLD of \textit{1.28}, FAD of \textit{0.85}, and Align Acc of \textit{97.73\%}. While DCF effectively captures temporal dependencies in the audio generation process, its reliance on static transformations within the latent space limits its ability to adapt to complex temporal variations in the video segments. This leads to suboptimal performance in metrics that evaluate semantic alignment (KLD) and audio quality (FAD).
GDCF extends DCF by incorporating additional flexibility for latent transformations, allowing it to dynamically adapt to the evolving temporal and semantic characteristics of the video. This results in substantial improvements across all metrics, with a KLD of \textit{0.97}, FAD of \textit{0.52}, and Align Acc of \textit{98.97\%}. The ability to model complex temporal dynamics enables GDCF to produce audio outputs that are not only temporally synchronized but also semantically richer and of higher quality compared to DCF.
GDCF achieves a significantly lower KLD than DCF (\textit{0.97} vs. \textit{1.28}), demonstrating superior semantic alignment between the generated audio and video inputs.  
The FAD of GDCF (\textit{0.52}) is markedly better than that of DCF (\textit{0.85}), indicating that GDCF generates audio outputs that are closer in distribution to real-world audio.  
GDCF also outperforms DCF in Align Acc, achieving near-perfect synchronization at \textit{98.97\%}, compared to DCF's \textit{97.73\%}. This reflects GDCF’s ability to handle fine-grained temporal dependencies effectively.
These results demonstrate the superiority of GDCF over DCF in balancing quality and efficiency. By extending variance-preserving flow with additional flexibility, GDCF captures both the distributional and temporal characteristics of video-audio tasks more effectively, leading to state-of-the-art performance across all evaluated metrics. This highlights the critical role of dynamic and adaptive flow-based objectives in achieving high-quality and synchronized video-to-audio generation.

\begin{table}[t]
\centering
\caption{Ablation studies on masking ratio for video-to-audio generation. The KLD, FAD, and Align Acc values are reported.}
\label{tab: ab_ratio}
\scalebox{0.76}{
\begin{tabular}{cccc}
\toprule
Masking Ratio & KLD $\downarrow$ & FAD $\downarrow$ & Align Acc $\uparrow$ \\ \midrule
0.3 & 1.25 & 1.06 & 98.12 \\
0.5 & 1.16 & 0.87 & 98.37 \\
0.7 & 1.09 & 0.65 & 98.78 \\
0.8 & \bf 0.97 & \bf 0.52 & \bf 98.97 \\
0.9 & 1.02 & 0.56 & 98.96 \\
\bottomrule
\end{tabular}}
\end{table}

\noindent\textbf{Impact of Masking Ratio.}
To analyze the effect of the masking ratio on the performance of the Video-Audio Masking Alignment (VAMA) module, we evaluate the model using masking ratios of \{0.3, 0.5, 0.7, 0.8, 0.9\}. The quantitative results are presented in Table~\ref{tab: ab_ratio}.
The masking ratio of 0.8 achieves the best results across all metrics, with a KLD of \textit{0.97}, FAD of \textit{0.52}, and Align Acc of \textit{98.97\%}. This demonstrates that the model effectively learns robust cross-modal dependencies and temporal synchronization when 80\% of the audio segments are masked. This ratio strikes the right balance between challenging the model to infer missing information and providing enough contextual cues for effective training.
When the masking ratio is reduced to 0.3 and 0.5, the performance degrades across all metrics. For instance, at a masking ratio of 0.3, the KLD increases to \textit{1.25}, FAD to \textit{1.06}, and Align Acc drops to \textit{98.12\%}. This indicates that lower masking ratios do not sufficiently challenge the model to learn robust representations, as most of the audio information is directly available, reducing the model's reliance on video guidance.
Increasing the masking ratio to 0.9 results in a slight performance drop, with a KLD of \textit{1.02}, FAD of \textit{0.56}, and Align Acc of \textit{98.96\%}. Although the degradation is minor, excessive masking likely leads to the loss of critical audio context, making it more difficult for the model to reconstruct missing segments accurately.
The results highlight the trade-offs between masking ratios. Lower ratios provide too much information, limiting the model's ability to learn meaningful dependencies, while higher ratios lead to excessive information loss. The optimal masking ratio of 0.8 balances these factors, encouraging the model to rely on cross-modal cues for reconstruction while retaining enough context for effective learning.
The masking ratio plays a crucial role in the effectiveness of the VAMA module. A ratio of 0.8 achieves the best performance across all metrics, enabling \method\ to generate audio that is semantically coherent, rhythmically aligned, and of high quality. This demonstrates the importance of selecting an appropriate masking ratio for training alignment models.

\section{Conclusion}

In this paper, we present \method, a novel framework for video-to-audio generation that achieves semantic and rhythmic coherence through innovative cross-modal masking and flow-based modeling techniques. 
Our \method\ addresses the limitations of previous approaches by introducing the Video-Audio Masking Alignment module to enforce fine-grained alignment between visual frames and audio segments, and the Generalized Video-Audio Flow (GVAF) module to enable efficient and high-quality audio generation.
Through extensive experiments on the VGGSound and AudioSet datasets, we demonstrated that \method\ achieves state-of-the-art performance across multiple metrics, including Kullback-Leibler Divergence (KLD), Fréchet Audio Distance (FAD), and Alignment Accuracy (Align Acc). Our framework outperforms existing methods by producing synchronized and contextually accurate audio outputs while significantly reducing inference time.
Our ablation studies further validated the effectiveness of the proposed components. 
Additionally, the choice of video and audio encoders, the use of Generalized Variance Preserving Flow, and an optimal masking ratio contributed to the success of \method.

{
    \small
    \bibliographystyle{ieeenat_fullname}
    \bibliography{reference}
}


\end{document}